\makeatletter
\def\input@path{{tex/}{./tex/}}
\makeatother

\documentclass[runningheads]{llncs}

\usepackage{accv}

\usepackage{accvabbrv}

\usepackage{graphicx}
\usepackage{booktabs}
\usepackage{colortbl}
\usepackage{amsmath}

\usepackage[accsupp]{axessibility}  

\newcommand{\PaperTitleEN}{FORUM: Frozen Outputs Reconciled Using Model Agreement for Visual Grounding}
\newcommand{\PaperTitleRunning}{FORUM for Visual Grounding}

\newcommand{\method}{FORUM}
\newcommand{\methodfull}{Frozen Outputs Reconciled \rev{Using Model Agreement} for Visual Grounding}

\newcommand{\TableOurs}{Ours}

\newif\ifrevmark
\revmarkfalse
\newcommand{\rev}[1]{\ifrevmark\textcolor{blue}{#1}\else#1\fi}
\newenvironment{revblock}{\ifrevmark\color{blue}\fi}{}
\DeclareCaptionFont{revblue}{\color{blue}}
\newcommand{\revcolor}{\ifrevmark\color{blue}\captionsetup{font+=revblue}\fi}

\definecolor{specgray}{gray}{0.5}

\definecolor{bandgray}{gray}{0.9}                       
\colorlet{oursblue}{blue!8}                              
\newcommand{\groupband}[2]{\rowcolor{bandgray}\multicolumn{#1}{l}{\textit{#2}}\\}

\DeclareMathOperator*{\argmax}{arg\,max}


\graphicspath{{figures/}}

\usepackage[breaklinks,colorlinks,citecolor=accvblue]{hyperref}

\usepackage{orcidlink}
\usepackage{pifont}   

\begin{document}

\title{\PaperTitleEN}

\titlerunning{\PaperTitleRunning}

\author{Taiyo Sato\inst{1,2}\thanks{Equal contribution.}\orcidlink{0009-0005-0477-2117}\textsuperscript{(\ding{41})} \and
Takamasa Sanda\inst{2}\protect\footnotemark[1] \and
Keisuke Maeda\inst{1}\orcidlink{0000-0001-8039-3462} \and
Takahiro Ogawa\inst{1}\orcidlink{0000-0001-5332-8112} \and
Miki Haseyama\inst{1}\orcidlink{0000-0003-1496-1761} \and
Shunya Nagashima\inst{2}\orcidlink{0009-0007-9741-7628}}
\authorrunning{T.~Sato et al.}
\institute{Hokkaido University, Sapporo, Japan\\
\email{\{taiyo\_sato,maeda,ogawa,mhaseyama\}@lmd.ist.hokudai.ac.jp}
\and
Neurogica Inc., Tokyo, Japan\\
\email{\{taiyo.sato,takamasa.sanda,shunya.nagashima\}@neurogica.com}}

\maketitle

\begingroup\renewcommand\thefootnote{}\footnotetext{\rev{Code is available at \url{https://github.com/Neurogica/FORUM}.}}\endgroup

\begin{abstract}
  Frozen multimodal large language models (MLLMs) now solve standard referring expression comprehension with a single prompted call, yet on adversarial benchmarks with same-category distractors and negation, even the largest models are confidently wrong, and resampling repeats the error.
  Models built from different data and architectures rarely fall for the same confounder, so their agreement is a strong label-free signal of the correct target.
  We present \method{}, a training-free test-time fusion of frozen MLLMs guided by two fixed geometric rules: agreement-based selection keeps the region supported by the most distinct models, and medoid localization returns an actual member box instead of a coordinate average, so one loose prediction cannot shift the answer.
  Fusing three open MLLMs, \method{} surpasses \rev{the} 397B-parameter \rev{published reference} by a relative 5\% in mean accuracy on the adversarial Ref-Adv-s benchmark, and a plain averaging ensemble by 15\%.
  The gains transfer to standard RefCOCO+, and a balanced lineup with no dominant member still surpasses the 397B model by 5\%.
\end{abstract}

\section{Introduction}
\label{sec:intro}

\begin{figure*}[t]
  \centering
  \includegraphics[width=0.92\linewidth]{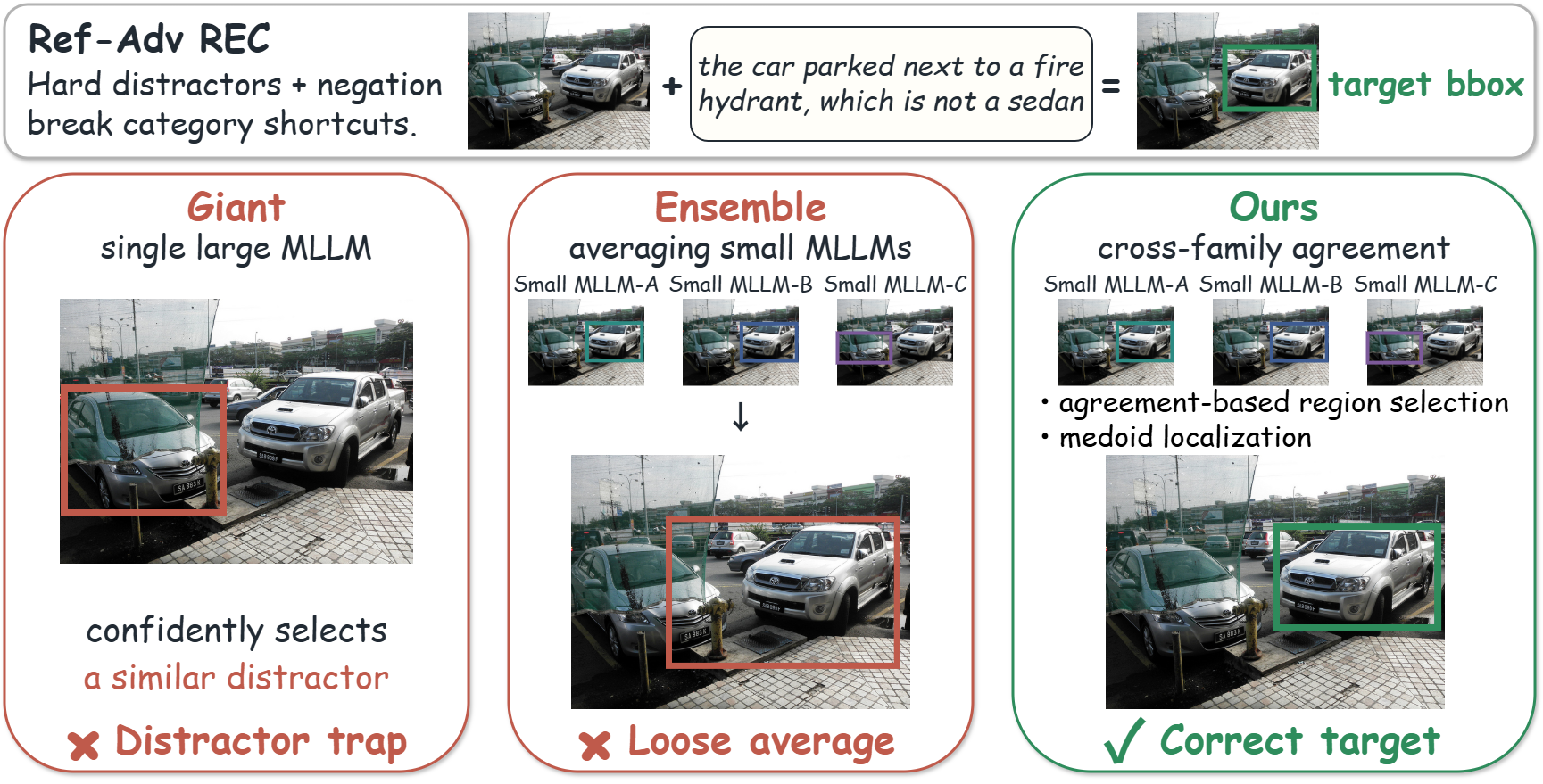}
  \caption{\textbf{\method{} fuses three \rev{different frozen} MLLMs at test time.}
  A single giant MLLM can confidently lock onto a look-alike distractor, and a plain
  ensemble that averages the predicted boxes drifts off-target. \method{} keeps the region
  where different frozen models agree and returns an actual member box, the \emph{medoid}.}
  \label{fig:teaser}
\end{figure*}

Referring expression comprehension (REC) asks a model to localize the object described by a natural-language expression in an image~\cite{kazemzadeh2014refcoco}.
Multimodal large language models (MLLMs) now reach strong scores on standard REC benchmarks simply by prompting a frozen model to output a bounding box~\cite{Bai2023QwenVL,Peng2024Kosmos2,Chen2023Shikra,You2024Ferret}, suggesting that task-specific detector pipelines are no longer required for many everyday grounding settings.

Recent adversarial benchmarks tell a different story.
Ref-Adv~\cite{dong2026refadv} pairs complex expressions with hard same-category distractors, so category shortcuts no longer suffice and models must reason over attributes, relations, and negation (\Cref{sec:problem}).
Even the largest published single models plateau on this split.
A single greedy grounding call can be confidently wrong when an MLLM locks onto a visually similar distractor, and resampling the same model tends to repeat the error rather than correct it~\cite{Rahmanzadehgervi2024Blind,dong2026refadv}.
Prior work improves grounding through specialized training or inference-time intervention~\cite{Jiang2026RexThinker,Liu2026AIF,Tao2026DiG,Zheng2026DeepEyes,Wu2026ForeSight}, but these directions typically change model weights, architecture, or the standard prompted-MLLM interface (\Cref{sec:related}).

We instead ask whether adversarial REC can be improved \emph{without} updating any weights, by fusing a small set of different frozen MLLMs at test time.
Our key observation is that errors made by sufficiently different models are often \rev{weakly correlated across families}: a look-alike distractor that traps one model need not trap another built from different data and architecture.
Self-consistency from a single model is therefore an unreliable correctness signal on adversarial inputs, whereas agreement across different models is much harder to fake when distractors are visually confounding.
At the same time, simply averaging the predicted boxes, as in a plain test-time ensemble, blends in loose or confidently wrong coordinates and collapses at strict IoU.
\rev{We show that test-time fusion of three frozen MLLMs surpasses a 397B-parameter single-model reference on Ref-Adv-s, with no fine-tuning (\Cref{sec:experiments}).}
\rev{The method is illustrated in \Cref{fig:teaser}.}

We propose \method{} (\methodfull{}), a training-free test-time fusion pipeline.
Three off-the-shelf MLLMs each contribute candidate boxes, and two fixed offline rules, region selection by cross-model agreement and localization without coordinate averaging, produce the final prediction (\Cref{sec:method}).

Our paper's contribution is outlined below:
\begin{itemize}
  \item We show that agreement among different frozen MLLMs \rev{provides a label-free, agreement-based confidence signal} on adversarial REC: their errors are \rev{weakly correlated across families (0.47 vs.\ 0.63 within-family)}, so the region on which independent models concur is more reliable than any single model's confidence, which cannot detect the distractor trap (\Cref{sec:method}).
  \item We propose medoid localization: rather than a coordinate average, we return the medoid, an actual member box, since under heterogeneous localization precision a single loose box drags an averaged box off-target and collapses strict-IoU accuracy (\Cref{sec:method}).
  \item Combining the two rules, a training-free fusion of three \rev{frozen} MLLMs surpasses \rev{the} $397$B-parameter \rev{published single-model reference} on Ref-Adv-s and improves over the strongest single member on RefCOCO+ (\Cref{sec:experiments}).
\end{itemize}

\section{Related Work}
\label{sec:related}

\subsection{Referring Expression Comprehension}

Referring expression comprehension (REC) localizes the image region described by a natural-language expression, and has been studied through a long line of benchmarks and methods reviewed in recent surveys~\cite{Qiao2021Survey,Xiao2025Survey}.
The RefCOCO and RefCOCO+ datasets~\cite{Yu2016Context}, together with RefCOCOg~\cite{Mao2016RefExp} and the earlier ReferItGame~\cite{kazemzadeh2014refcoco}, established the standard protocol of grounding expressions to bounding boxes on natural images.
Methods progressed from pairing object detectors with expression encoders to end-to-end vision-language detection such as MDETR~\cite{Kamath2021MDETR}, GLIP~\cite{Li2022GLIP}, and Grounding DINO~\cite{Liu2024GroundingDINO}, building on joint vision-language pretraining such as UNITER~\cite{Chen2020UNITER}.
A central difficulty is that the target is defined not only by category but also by context, attributes, and relations to \rev{other} objects.

Recent work revisits REC as a testbed for visual reasoning under harder distractors.
Ref-Adv~\cite{dong2026refadv} builds expressions with hard same-category distractors and linguistic facets such as negation; we describe the benchmark in \Cref{sec:problem}.
Some recent methods pursue stronger reasoning through training: for example, Rex-Thinker casts object referring as candidate-wise chain-of-thought reasoning, identifying candidate instances and verifying whether each satisfies the expression~\cite{Jiang2026RexThinker}.
In contrast, we study whether different frozen MLLMs can correct one another at test time when their distractor errors are \rev{weakly correlated across families}.

\subsection{Multimodal Large Language Models for Visual Grounding}

General-purpose multimodal large language models (MLLMs) have become strong grounding baselines.
Qwen-VL~\cite{Bai2023QwenVL} couples localization, text reading, and instruction following in a single model family, so that a frozen MLLM can be prompted to output bounding boxes directly, and grounded MLLMs such as Kosmos-2~\cite{Peng2024Kosmos2}, Shikra~\cite{Chen2023Shikra}, and Ferret~\cite{You2024Ferret} add explicit mechanisms linking text spans, coordinates, and image regions.
These models show that box grounding can be handled by MLLMs without task-specific detector pipelines, which motivates evaluating pure MLLM baselines on adversarial REC.

Despite this progress, MLLMs can still struggle with precise spatial perception: they perform well on high-level image-text benchmarks yet fail on simple tasks requiring exact spatial information, especially when objects are close, overlapping, or small~\cite{Rahmanzadehgervi2024Blind}.
High-resolution designs mitigate a related issue by increasing visual detail through image splitting or dynamic resolution~\cite{Xu2024LLaVAUHD}, and recent grounding-oriented methods control inference-time information flow, strengthen fine-grained perception, or encourage models to revisit image evidence during reasoning~\cite{Liu2026AIF,Tao2026DiG,Zheng2026DeepEyes,Wu2026ForeSight}.
These approaches improve MLLM perception, but they typically require specialized training, internal model intervention, or tool-style reasoning beyond a standard prompted MLLM.

\subsection{Test-Time Fusion and Ensembling}

Combining several predictions at test time is a classic way to improve reliability without changing weights.
For object detection, Soft-NMS~\cite{Bodla2017SoftNMS} and weighted boxes fusion~\cite{Solovyev2021WBF} aggregate many overlapping detections, but assume calibrated detector scores and dense proposals rather than a few prompted MLLM boxes.
Training-free model merging such as model soups~\cite{Wortsman2022ModelSoups} and test-time augmentation~\cite{Shanmugam2021TTA} combine weights or augmented views of a single model, not the outputs of different models.
For language models, agreement across samples or models improves answers: self-consistency aggregates multiple reasoning paths~\cite{Wang2023SelfConsistency}, multi-agent debate reconciles independent answers~\cite{Du2024Debate}, and inference-time fusion of several LLMs combines their generations~\cite{Jiang2023LLMBlender,Li2024MoreAgents}.

Our setting differs from these lines of work in what signal drives selection and how the final box is formed.
First, Soft-NMS~\cite{Bodla2017SoftNMS} and weighted boxes fusion~\cite{Solovyev2021WBF} require calibrated per-detector confidence to suppress or weight candidates, but a greedy MLLM call exposes no such score, and \Cref{sec:exp-analysis} shows that even where a confidence proxy is available, it is exactly the signal that fails on the distractor trap; we therefore select without confidence rather than approximate one.
We count agreement by \emph{model identity} (\Cref{sec:method-agreement}) rather than candidate density, which is the criterion classical spatial-consensus clustering optimizes for.
Second, having selected a region, we return an actual member box rather than fitting a coordinate average or a weighted estimator to it, so a single imprecise member can only be excluded, never blended in.
\method{} keeps the prompted-MLLM interface, uses no training, and studies when \rev{weakly correlated} errors beat both single-model scaling and plain averaging.

\section{Problem Formulation}
\label{sec:problem}

We study referring expression comprehension on the adversarial Ref-Adv benchmark~\cite{dong2026refadv}.

\subsection{Task and Metric}
\label{sec:rec-task}

Referring expression comprehension (REC) localizes a target object in an image from a natural-language description~\cite{kazemzadeh2014refcoco}.
Each instance consists of an RGB image~$I$ and a referring expression~$e$, and the model must output a single axis-aligned bounding box $b=(x_1,y_1,x_2,y_2)$ for the object uniquely described by~$e$.
Let $b^\ast$ denote the ground-truth box.
We measure localization with the intersection-over-union $\mathrm{IoU}(b,b^\ast)=|b\cap b^\ast|/|b\cup b^\ast|$, and a prediction is correct at threshold~$\tau$ when $\mathrm{IoU}(b,b^\ast)\ge\tau$.
We report accuracy $\mathrm{Acc}@\tau$ at several thresholds; a prediction that does not parse into a valid box receives zero accuracy under the official scorer.

\subsection{The Ref-Adv Benchmark}
\label{sec:refadv}

Standard REC datasets such as RefCOCO often let models succeed via shortcuts (\eg, the category name alone) without fine-grained visual reasoning.
Ref-Adv~\cite{dong2026refadv} is an adversarial benchmark that reduces such shortcuts by pairing complex expressions with hard visual distractors: objects of the same category that do not satisfy the linguistic constraints in~$e$.
On average it has long expressions ($11.5$ words), several distractors per image ($4.01$, at least two per case), and a substantial negation ratio ($21.25\%$), with images drawn from COCO val2017 and OpenImages~\cite{dong2026refadv}.
We evaluate on the public subset \emph{Ref-Adv-s}, an evaluation-only split of $1{,}142$ cases ($40\%$ negated) with released scoring code and reference predictions.

\subsection{Evaluation Protocol}
\label{sec:eval-protocol}

We score every method with the unmodified official Ref-Adv harness~\cite{dong2026refadv}, on the full Ref-Adv-s split.
Following Ref-Adv, the primary metric is the mean of $\mathrm{Acc}@\tau$ over $\tau\in\{0.5,0.75,0.9\}$, and we also report each threshold separately, because a loose box that passes a lenient threshold can still overlap an adjacent distractor at a strict one.
We further analyze performance by distractor count and by the presence of negation.
We focus on single-target REC, where each instance has exactly one ground-truth box; segmentation, multi-box prediction, and video grounding are out of scope.

\section{Method}
\label{sec:method}

\begin{figure*}[t]
  \centering
  \includegraphics[width=\linewidth]{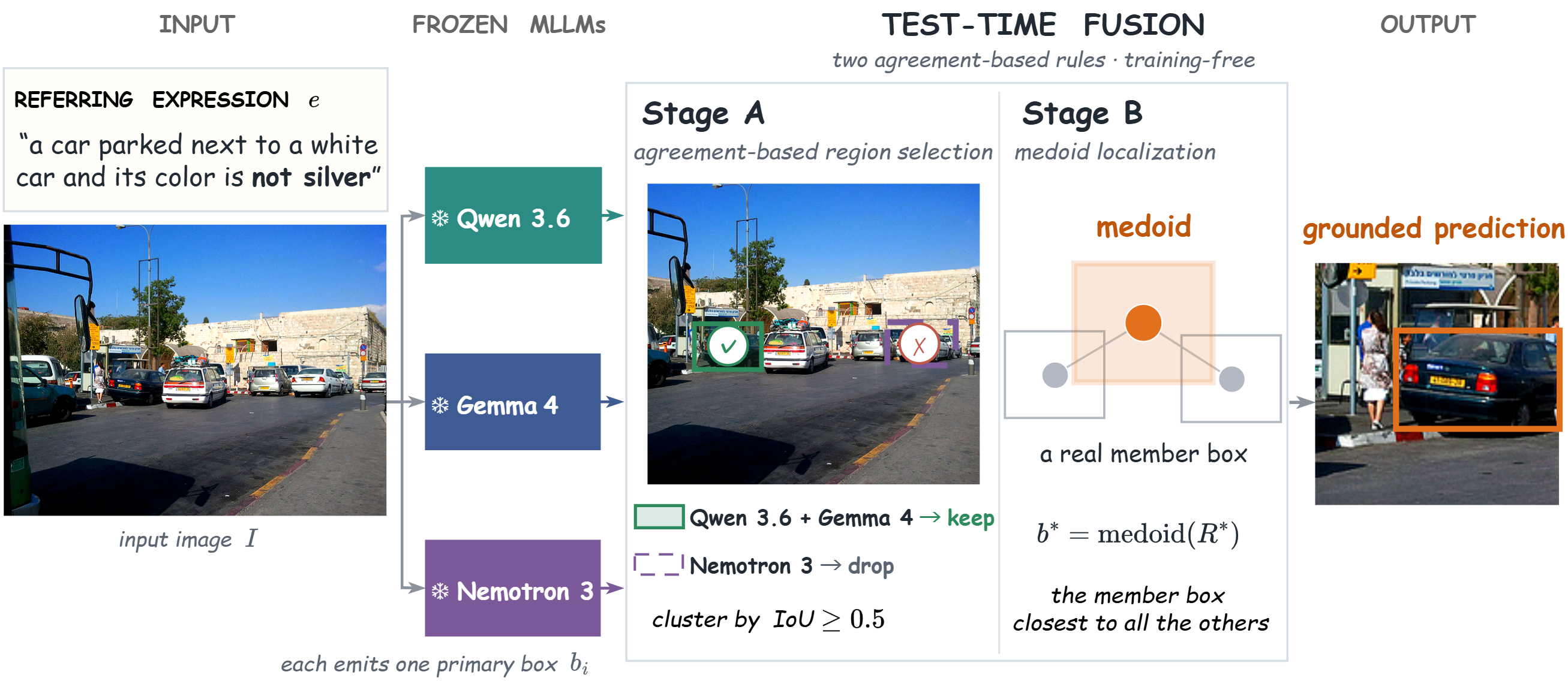}
  \caption{\textbf{\rev{Agreement-based} test-time fusion of heterogeneous MLLMs.}
  The image~$I$ and expression~$e$ are sent to the three frozen MLLMs of the dominant
  lineup, each emitting one primary box; for the query (\emph{``\dots its color is not
  silver''}), Nemotron is trapped on the silver car while Qwen3.6 and Gemma-4 agree on the
  correct one.
  \textbf{Stage~A} clusters the boxes by IoU and keeps the region supported by the most
  distinct models ($\checkmark$), discarding the outlier ($\times$).
  \textbf{Stage~B} outputs the \emph{medoid}, an actual member box, rather than a synthetic
  average that a loose box could drag off-target.}
  \label{fig:method}
\end{figure*}

We propose \method{} (\methodfull{}), a \emph{training-free} test-time fusion of different \rev{MLLMs} for adversarial \rev{REC}.
We query $K$ frozen MLLMs from different model families, pool the boxes they emit, and apply two fixed offline rules, agreement-based region selection and medoid localization, to produce \rev{the output box}~$b$ without fine-tuning or coordinate averaging.

\subsection{Overview}
\label{sec:method-overview}

Ref-Adv instances contain several same-category distractors, so a single greedy grounding call can be confidently wrong.
Self-consistency from one model cannot detect this \emph{distractor trap} in principle: a model locked onto a look-alike may repeat the same error under resampling.
We instead exploit \emph{cross-model agreement}, since models built from different data and architectures rarely fall for the same confounder at once.
Our pipeline (\Cref{fig:method}) has two stages.
Stage~A performs agreement-based region selection: it clusters the candidate boxes and keeps the region supported by the largest number of distinct models.
Stage~B performs medoid localization: it returns an actual member box of the selected cluster rather than an averaged coordinate, so a single loose box cannot drag the answer off-target.

Three constraints shape this design.
The fusion must be training-free and must not require weight or logit access, so that any prompted model, hosted or open, can serve as a member.
It must add no model calls beyond the members themselves, keeping the cost at $K$ greedy forwards per instance.
And it must not depend on calibrated confidences, which are exactly the signal that adversarial distractors defeat (\Cref{sec:exp-analysis}).
The two rules below satisfy all three: they \rev{use} only the $K$ output boxes \rev{and a single IoU threshold}.

\subsection{Cross-Model Agreement Selection}
\label{sec:method-agreement}

For each instance, model~$i$ emits a primary box at temperature~$0$ and may optionally contribute further candidates~$C_i$, with the primary box included in~$C_i$.
We pool all candidates into a set $\mathcal{C}=\bigcup_i C_i$, where each candidate $c\in\mathcal{C}$ carries its source model $m(c)$.
We form clusters as the connected components of the graph in which two candidates are linked when their IoU is at least a threshold~$\tau$ (single-link clustering).
The threshold is intentionally permissive, because boxes from different models can differ in scale even when they refer to the same object; the medoid step below still returns an actual member box, which limits any chaining effect.

Among the clusters~$\mathcal{R}$, we keep the one supported by the most distinct models, breaking ties by cluster size:
\begin{equation}
  \mathcal{R}^{\dagger}=\argmax_{R\in\mathcal{R}}\;\bigl|\{\,m(c):c\in R\,\}\bigr|,
  \qquad
  R^{\ast}=\argmax_{R\in\mathcal{R}^{\dagger}}\;|R|,
  \label{eq:agreement-select}
\end{equation}
where $|\{\,m(c):c\in R\,\}|$ counts the distinct models present in cluster~$R$.
\rev{The pipeline is deterministic: members run at temperature~$0$ in a fixed order, a remaining cluster tie goes to the cluster containing the earliest candidate, and a medoid tie in \Cref{sec:method-medoid} goes to the earliest box.}
Intuitively, \Cref{eq:agreement-select} discards spatial outliers and keeps the region where independent models agree.
The rule counts \emph{distinct models} rather than raw candidates, so a member that contributes several near-duplicate candidates cannot outvote the others.
Agreement is informative because errors are \rev{weakly correlated across families}: members trained on different data with different architectures rarely prefer the same wrong object, so a region supported by several of them is unlikely to be a shared mistake, whereas the correct object attracts every member that resolves the expression.
This is a measured property of the lineup, not an assumption: cross-family error correlation is substantially lower than within-family correlation (0.47 vs.\ 0.63; \Cref{sec:exp-analysis}).
Empirically, correctness rises monotonically with the number of agreeing models, so agreement acts as a \rev{label-free, agreement-based confidence signal} under hard distractors (\Cref{sec:exp-analysis}).

\subsection{Medoid Localization}
\label{sec:method-medoid}

A plain ensemble averages the boxes in the selected cluster, which fails on adversarial data when one member draws a loose box: averaging pulls the prediction toward the background and collapses strict IoU (\Cref{sec:exp-ablation}).
Given the selected set~$R^{\ast}$, we instead output the \emph{medoid}, the member box that overlaps the rest of the cluster most:
\begin{equation}
  b^{\ast}=\argmax_{c\in R^{\ast}}\;\sum_{c'\in R^{\ast}}\mathrm{IoU}(c,c'),
  \label{eq:medoid}
\end{equation}
\ie, the medoid of~$R^{\ast}$ under the box dissimilarity $1-\mathrm{IoU}$.
Because $b^{\ast}$ is an existing member prediction, imprecise boxes are excluded automatically when they are spatial outliers, without labeling any model as weak.
Under heterogeneous localization precision, this realizes the intuition of precision-weighted fusion (Gauss--Markov / BLUE) while never producing a synthetic box.

The choice among aggregation rules follows from what each assumes.
Averaging is justified only when member boxes are equally precise; inverse-variance weighting is the optimal linear rule when precisions differ, but it must estimate them and, like averaging, produces a synthetic box that no member drew.
\rev{The inverse-variance baseline used in \Cref{sec:exp-ablation} does not estimate a per-box variance: each member receives one global weight proportional to its Acc@0.75, and the selected boxes are combined as a precision-weighted mean.}
\rev{Coordinate-wise median is another synthetic-box rule and inherits coordinate-wise robustness, yet it still blends boxes that no single member emitted.}
The medoid needs no precision estimates and \rev{returns an existing member prediction}: an arbitrarily loose member changes the output only by switching which real box is selected, never by dragging coordinates continuously off the object.
\Cref{sec:exp-ablation} confirms this empirically: with near-equal members \rev{averaging and inverse-variance coincide, median and medoid stay close}, and once a coarse member joins, averaging collapses at strict IoU while the medoid is unaffected.

\section{Experiments}
\label{sec:experiments}

\subsection{Experimental Setup}
\label{sec:exp-setup}

\paragraph{Benchmarks and metrics.}
\rev{Our primary evaluation is Ref-Adv-s ($n{=}1{,}142$; \Cref{sec:refadv}) under the official harness and metrics (\Cref{sec:eval-protocol}); the standard RefCOCO+ splits (val, testA, testB) test whether gains transfer beyond the adversarial regime (\Cref{sec:exp-standard}).}
\rev{Alongside the mean, we regard the strict \emph{Acc@0.9} as the most informative signal, since only precise localization confirms that the correct object, rather than a look-alike, was grounded.}
All methods are training-free: model weights stay frozen and fusion uses fixed rules ($\tau{=}0.5$, medoid localization; \Cref{sec:method}) with no test-set calibration of selection parameters; mean accuracy varies by at most $0.4$ on Ref-Adv-s and $1.0$ on RefCOCO+ testB across $\tau\in[0.3,0.7]$, so the result is not tuned to $\tau{=}0.5$ (Appendix A).

\paragraph{Fusion lineups.}
Our base lineup fuses three off-the-shelf MLLMs from different families: Qwen3.6-35B-A3B~\cite{QwenTeam2026Qwen36A3B}, Gemma-4-31B~\cite{Google2026Gemma4}, and Nemotron-3-Nano-Omni-30B-A3B~\cite{NVIDIA2025Nemotron} (two 3B-active MoE models and one dense 31B model).
We call it the \emph{dominant} lineup, since Qwen3.6 is by far its strongest member.
To show that the gain does not depend on one strong model, we also evaluate a \emph{balanced} lineup of near-equal members: Qwen3.5-27B~\cite{QwenTeam2026Qwen35}, Gemma-4-31B, and GLM-4.6V-Flash~\cite{GLMVTeam2025GLM45V} (GLM-4.6V for short), dense models of 27B, 31B, and roughly 10B parameters whose single-model means span only 1.6 points.
\rev{Both lineups follow one selection rule: a candidate must have open weights and output boxes under its documented grounding prompt, its single-model quality is screened on RefCOCOg val, and a lineup spans distinct base families and size bands.}
\rev{Of eight screened candidates, three were excluded on family and size band, not on quality.}
\rev{Ref-Adv-s single-model accuracy only labeled a lineup dominant or balanced; no lineup was chosen by fused accuracy.}
\TableOurs{} applies to any set of sufficiently different models\rev{, by which we mean members from distinct base families (cross-family error correlation 0.47 vs.\ 0.63 within family; \Cref{sec:exp-analysis})}.
Each member emits one greedy primary box at temperature~$0$; for the agreement-versus-confidence analysis we additionally draw $k{=}8$ stochastic samples from the balanced lineup's Qwen member (Appendix B).

\paragraph{Compared systems.}
We compare (i)~\emph{single-model} greedy grounding for each fusion member and published leaderboard entries;
(ii)~\emph{plain ensemble}, which averages coordinates in the largest IoU cluster (standard test-time box-ensemble baseline); and
(iii)~\TableOurs{}, cross-model agreement selection and medoid localization.
The single-MLLM baselines cover Qwen, InternVL3.5, GLM-4.6V,
Youtu-VL, Gemma~4, and Nemotron families~\cite{
dong2026refadv,Bai2025Qwen25VL,Bai2025Qwen3VL,
QwenTeam2026Qwen35,QwenTeam2026Qwen36A3B,
Wang2025InternVL35,GLMVTeam2025GLM45V,Wei2026YoutuVL,
Google2026Gemma4,NVIDIA2025Nemotron}.
Published leaderboard numbers are cited as reference points only; our runs use the same harness and coordinate conventions, and the single rows for fusion members are our greedy runs (for Qwen3.5-27B the published leaderboard entry is 51.6).
\Cref{tab:leaderboard} shows the fusion members, the strongest published single model, and the specialist baselines; the full comparison is in Appendix E.

\begin{table}[tb]
  \centering
  \caption{Ref-Adv-s results (mean of Acc@\{0.5, 0.75, 0.9\}), sorted by mean within each group; best per column in bold. Fusion rows use the dominant lineup; the full comparison including the balanced lineup is in Appendix E.}
  \label{tab:leaderboard}
  \small
  \setlength{\aboverulesep}{0pt}\setlength{\belowrulesep}{0pt}
  \renewcommand{\arraystretch}{1.15}
  \setlength{\tabcolsep}{3.5pt}
  \begin{tabular}{lcccc}
    \toprule
    System & Mean\,$\uparrow$ & @0.5\,$\uparrow$ & @0.75\,$\uparrow$ & @0.9\,$\uparrow$ \\
    \midrule
    \groupband{5}{Specialist}
    Grounding DINO~\cite{Liu2024GroundingDINO} & 15.9 & 19.7 & 17.2 & 10.7 \\
    DeRIS-B~\cite{Dai2025DeRIS} & 21.1 & 28.3 & 22.2 & 12.8 \\
    \midrule
    \groupband{5}{MLLMs}
    Nemotron-Omni-30B & 21.5 & 51.5 & 9.5 & 3.6 \\
    GLM-4.6V & 48.3 & 60.0 & 51.4 & 33.5 \\
    Gemma-4-31B & 49.4 & 64.7 & 53.5 & 30.0 \\
    Qwen3.5-27B & 49.9 & 64.3 & 52.4 & 32.9 \\
    Qwen3.6-35B-A3B & 52.2 & 65.1 & 55.3 & 36.3 \\
    Qwen3.5-397B-A17B-FP8 & 52.6 & 68.0 & 55.6 & 34.2 \\
    \midrule
    \groupband{5}{Test-time fusion}
    Plain ensemble & 48.2 & \textbf{70.2} & 57.7 & 16.7 \\
    \rowcolor{oursblue}
    \TableOurs{} & \textbf{55.4} & 69.8 & \textbf{59.2} & \textbf{37.2} \\
    \bottomrule
  \end{tabular}
\end{table}

\subsection{Implementation Details}
\label{sec:exp-impl}

Each MLLM uses its family-specific direct grounding prompt.
Fusion pools the greedy primary boxes, optionally augmented with stochastic samples; the two selection stages add no further model calls.
Per instance, the cost is three greedy forwards (one per member) plus any optional sampling passes; per-forward active sizes are 3B, 31B, and 3B for the dominant lineup and 27B, 31B, and roughly 10B (dense) for the balanced one.
\rev{Measured on one 8$\times$H100 node (Appendix H), FLOPs favor the 397B reference (19.6 vs.\ 29.7 TFLOPs per instance for the dominant lineup), whereas memory favors \method{}, whose largest member fits one GPU (67 vs.\ 378\,GiB); we therefore make no efficiency claim.}

\subsection{Main Results on Ref-Adv-s}
\label{sec:exp-main}

\Cref{tab:leaderboard} summarizes Ref-Adv-s performance.
Despite using no training, \TableOurs{} reaches \textbf{55.4 mean} with the dominant lineup and 55.3 with the balanced one, exceeding every published single-model entry up to the 397B Qwen3.5-A17B reference (52.6 mean), with the best strict-IoU performance (37.2 Acc@0.9).
Dedicated REC specialists fare far worse on this adversarial split: an open-set grounding detector (Grounding DINO~\cite{Liu2024GroundingDINO}, full expression) and a referring-segmentation model (DeRIS~\cite{Dai2025DeRIS}) reach only 15.9 and 21.1 mean, as they are not built for the complex, often negated language in Ref-Adv.
The plain ensemble exposes the localization failure mode: it is slightly higher at the lenient Acc@0.5 but collapses at the strict Acc@0.9 (16.7 vs.\ 37.2), because the loose Nemotron boxes drag the average off-target.
Averaging the balanced lineup is instead competitive (54.5; \Cref{tab:localize-ablation}), above even the 397B reference, since most of that lineup's gain comes from Stage-A selection over \rev{weakly correlated} members; Stage~B's contribution there is a smaller, targeted correction at strict IoU that becomes the dominant factor precisely when a member is imprecise, as on the dominant lineup (\Cref{sec:exp-ablation}).
In both cases \TableOurs{} is better.
A paired bootstrap over the full split ($10{,}000$ resamples) confirms the gains: $+3.2$ mean over the best member and $+7.2$ over the plain ensemble, both $p<0.001$.
The balanced lineup's margins are likewise significant; full confidence intervals are in Appendix C.

\subsection{Transfer to Standard REC}
\label{sec:exp-standard}

\Cref{tab:refcoco-plus} extends the evaluation to the standard RefCOCO+ splits (val, testA, testB) with the dominant lineup, with Ref-Adv-s shown as the adversarial reference.
RefCOCO+ is the hard case for fusion: Qwen3.6 is by far the strongest member there, so fusion must not fall below its dominant member.
\TableOurs{} is the best system on RefCOCO+ val (76.4 mean) and testB (69.2), and is within $0.1$ of the strongest single member on testA (80.3 vs.\ 80.4 for Qwen3.6), while beating the plain ensemble by a wide margin on every split.
\rev{At Acc@0.5, \TableOurs{} reaches 89.0/92.5/81.5 on val/testA/testB against 87.8/91.8/79.3 for Qwen3.6, above the strongest member on every split.}
The gains therefore transfer beyond the adversarial regime and hold even when one member dominates.

\begin{table}[tb]
  \centering
  \caption{Transfer across REC splits with the dominant lineup (mean of Acc@\{0.5, 0.75, 0.9\}); R+ denotes RefCOCO+. Best per column in bold.}
  \label{tab:refcoco-plus}
  \setlength{\aboverulesep}{0pt}\setlength{\belowrulesep}{0pt}
  \renewcommand{\arraystretch}{1.15}
  \setlength{\tabcolsep}{5pt}
  \begin{tabular}{lcccc}
    \toprule
    System & Ref-Adv-s\,$\uparrow$ & R+ val\,$\uparrow$ & R+ testA\,$\uparrow$ & R+ testB\,$\uparrow$ \\
    \midrule
    Nemotron (single) & 21.5 & 31.9 & 34.0 & 27.2 \\
    Plain ensemble & 48.2 & 60.8 & 64.3 & 55.7 \\
    Gemma-4 (single) & 49.4 & 61.8 & 59.2 & 58.8 \\
    Qwen3.6 (single) & 52.2 & 75.5 & \textbf{80.4} & 67.5 \\
    \rowcolor{oursblue}
    \TableOurs{} & \textbf{55.4} & \textbf{76.4} & 80.3 & \textbf{69.2} \\
    \bottomrule
  \end{tabular}
\end{table}

\begin{figure}[t]
  \centering
  \includegraphics[width=0.80\linewidth]{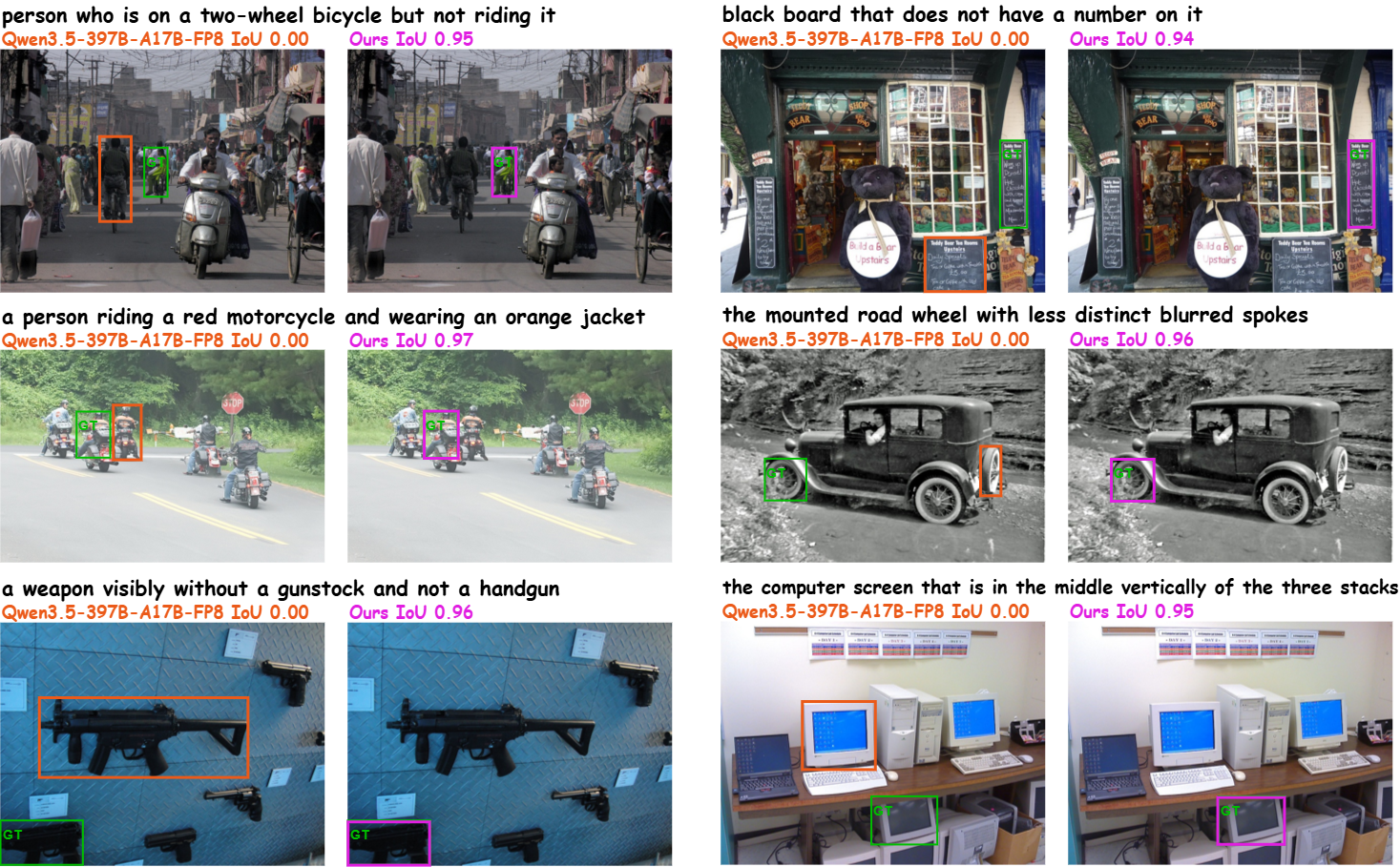}
  \caption{Successful Ref-Adv-s examples.}
  \label{fig:qual-refadv-success}
\end{figure}

\begin{figure}[t]
  \centering
  \includegraphics[width=0.80\linewidth]{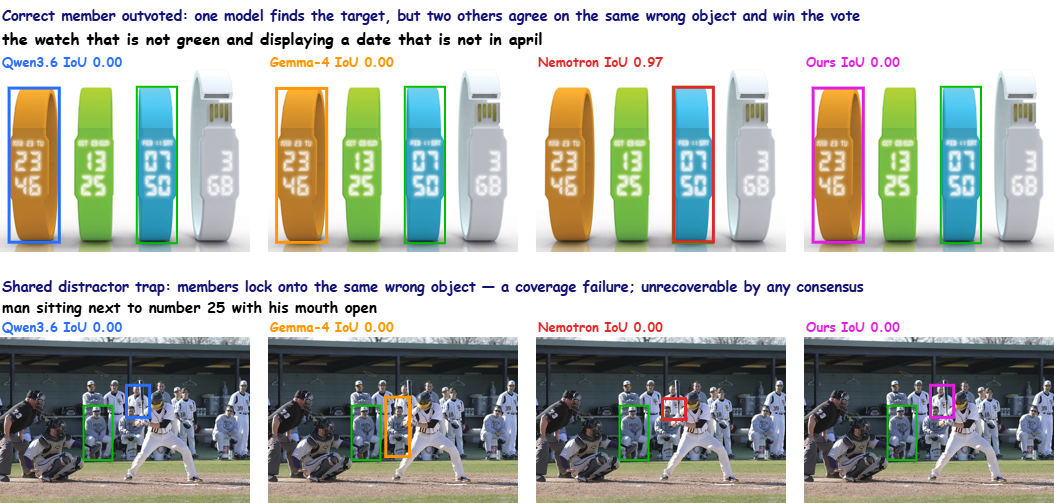}
  \caption{Failure modes on Ref-Adv-s.}
  \label{fig:qual-refadv-failure}
\end{figure}

\subsection{Qualitative Results and Failure Analysis}
\label{sec:exp-qualitative}

\Cref{fig:qual-refadv-success,fig:qual-refadv-failure} show representative Ref-Adv-s cases for the dominant lineup; additional examples are in Appendix F.
In successful cases, \TableOurs{} suppresses same-category distractors by selecting the region supported by multiple model families, even when an individual model proposes a plausible but linguistically inconsistent object.

\begin{figure}[t]
  \centering
  \includegraphics[width=0.60\linewidth]{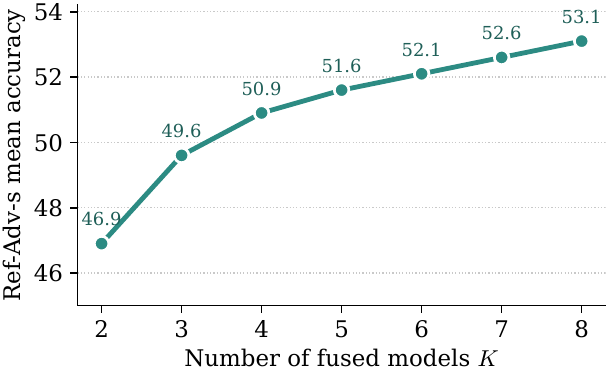}
  \caption{Scaling the number of fused small ($\le$10B) models on Ref-Adv-s: mean accuracy averaged over all $\binom{8}{K}$ size-$K$ subsets of an eight-model pool rises monotonically with~$K$.}
  \label{fig:kscaling}
\end{figure}

\rev{A quantitative decomposition characterizes the failures in \Cref{fig:qual-refadv-failure}.}
On the dominant lineup, \TableOurs{} fails at Acc@0.5 on 30\% of instances.
In 72\% of these failures no member localizes the target at all, so no selection rule could recover them: coverage, not selection, is the dominant bottleneck, and it is exactly the part that adding a further \rev{weakly correlated} member improves (\Cref{fig:kscaling}).
In the remaining 28\% a correct member box exists but is not chosen, the price of trusting agreement over any single member.
An oracle that always picked the best member box would reach 78.2 Acc@0.5; between the strongest single member (65.1) and this ceiling, \TableOurs{} recovers 36\% of the headroom without labels, confidences, or extra model calls.
The two cases in \Cref{fig:qual-refadv-failure} are instances of these two categories: shared distractors are coverage failures, and outvoting is a selection failure, so the qualitative and quantitative pictures agree.

\begin{table}[tb]
  \centering
  \caption{Ref-Adv-s mean accuracy ($\uparrow$) by facet, dominant lineup ($n$ in parentheses). Best per column in bold.}
  \label{tab:facet}
  \small
  \setlength{\aboverulesep}{0pt}\setlength{\belowrulesep}{0pt}
  \renewcommand{\arraystretch}{1.15}
  \setlength{\tabcolsep}{5pt}
  \begin{tabular}{lccccc}
    \toprule
     & \multicolumn{2}{c}{Negation} & \multicolumn{3}{c}{Distractors} \\
    \cmidrule(lr){2-3}\cmidrule(lr){4-6}
    System & yes & no & 2--3 & 4--6 & $\ge$7 \\
     & {\scriptsize(457)} & {\scriptsize(685)} & {\scriptsize(698)} & {\scriptsize(315)} & {\scriptsize(129)} \\
    \midrule
    Qwen3.6 (single) & 57.5 & 48.7 & 50.9 & 54.5 & 54.3 \\
    Plain ensemble & 52.7 & 45.3 & 48.9 & 47.7 & 45.5 \\
    \rowcolor{oursblue}
    \TableOurs{} & \textbf{61.2} & \textbf{51.5} & \textbf{55.2} & \textbf{56.3} & \textbf{54.5} \\
    \bottomrule
  \end{tabular}
\end{table}

\subsection{Agreement Analysis and Facets}
\label{sec:exp-analysis}

Cross-model agreement, not per-model confidence, is the reliable selection signal.
Across an eight-model pool, cross-family error correlation (0.47) is well below within-family correlation (0.63), and correctness rises monotonically with the number of agreeing members ($P(Y \mid a{=}1,2,3) = 31\%, 60\%, 85\%$ on the balanced lineup), so agreement acts as \rev{an agreement-based confidence signal} that \rev{per-model} confidence cannot replace (\Cref{sec:method-agreement}).
\rev{Matched controls (Appendix G) separate this from single-model strength and budget: cross-family fusion reaches 54.6 from an individual mean of 49.2 ($+5.4$), a same-family Qwen trio 52.5 from a higher 50.0 ($+2.5$), with error correlations 0.47 and 0.62, and three samples of the strongest member reach 50.4.}
A model that is confident under resampling yet contradicted by the other members is right only 33.3\% of the time; the full confidence-versus-agreement analysis is in Appendix B.

\paragraph{Scaling the number of fused models.}
To test whether the benefit grows with the number of members, we draw subsets from a pool of eight small ($\le$10B) models spanning five families (Qwen3.5-4B/9B, Qwen3-VL-8B, Gemma-3n-E4B, InternVL3.5-4B/8B, Youtu-VL, GLM-4.6V) and fuse each subset with the same rules.
\Cref{fig:kscaling} reports mean accuracy averaged over \emph{all} $\binom{8}{K}$ size-$K$ subsets, so the trend is order-independent and involves no model selection: accuracy rises monotonically with~$K$, consistent with the agreement mechanism of \Cref{sec:method-agreement}.

\paragraph{Robustness across facets.}
\Cref{tab:facet} breaks Ref-Adv-s down by linguistic and visual difficulty for the dominant lineup.
\TableOurs{} improves over both the best single member and the plain ensemble on every facet.
The gain over the single model is largest on negated expressions ($+3.7$ mean) and few-distractor scenes, and narrows on the hardest scenes with seven or more distractors, where little \rev{complementary} signal remains.

\begin{table}[tb]
  \centering
  \caption{Localization ablation on Ref-Adv-s (same agreement-selected clusters) for both lineups. Only the medoid is an actual member prediction; best per column and band in bold.}
  \label{tab:localize-ablation}
  \small
  \setlength{\aboverulesep}{0pt}\setlength{\belowrulesep}{0pt}
  \renewcommand{\arraystretch}{1.15}
  \setlength{\tabcolsep}{3.5pt}
  \begin{tabular}{lcccc}
    \toprule
    Final box rule & Mean\,$\uparrow$ & @0.5\,$\uparrow$ & @0.75\,$\uparrow$ & @0.9\,$\uparrow$ \\
    \midrule
    \groupband{5}{Dominant lineup}
    Coordinate average (plain ensemble) & 48.2 & 70.2 & 57.7 & 16.7 \\
    Inverse-variance weighted & 55.0 & \textbf{70.7} & \textbf{60.7} & 33.5 \\
    \rev{Coordinate-wise median} & \textbf{\rev{55.5}} & \rev{70.3} & \rev{59.5} & \rev{36.7} \\
    \rowcolor{oursblue}
    Medoid (\TableOurs{}) & 55.4 & 69.8 & 59.2 & \textbf{37.2} \\
    \midrule
    \groupband{5}{Balanced lineup}
    Coordinate average (plain ensemble) & 54.5 & \textbf{69.4} & 59.5 & 34.5 \\
    Inverse-variance weighted & 54.5 & \textbf{69.4} & 59.5 & 34.5 \\
    \rev{Coordinate-wise median} & \textbf{\rev{55.7}} & \rev{69.2} & \textbf{\rev{59.6}} & \textbf{\rev{38.4}} \\
    \rowcolor{oursblue}
    Medoid (\TableOurs{}) & 55.3 & 69.1 & 59.5 & 37.2 \\
    \bottomrule
  \end{tabular}
\end{table}

\subsection{Ablations}
\label{sec:exp-ablation}

\paragraph{Localization rule.}
\Cref{tab:localize-ablation} isolates the final-box rule on the same agreement-selected clusters, for both lineups.
On the dominant lineup, whose Nemotron member draws notoriously loose boxes (9.5 Acc@0.75 as a single; \Cref{tab:leaderboard}), averaging collapses at strict IoU (16.7 Acc@0.9) while the medoid barely moves (37.2); inverse-variance weighting recovers most of the gap but, like averaging, still returns a \emph{synthetic} box.
\rev{Coordinate-wise median stays close to the medoid there (55.5 mean, 36.7 Acc@0.9 vs 55.4 and 37.2) and is also a synthetic box.}
On the balanced lineup, member precisions are near-equal, so inverse-variance weighting degenerates to plain averaging (identical to one decimal at every threshold), as the Gauss--Markov view in \Cref{sec:method-medoid} predicts, and the localization gain \rev{of the medoid over averaging} is small but significant ($+0.8$ mean, $+2.7$ Acc@0.9; Appendix C).
\rev{The coordinate-wise median is slightly ahead of the medoid on that lineup (55.7 mean, 38.4 Acc@0.9 vs 55.3 and 37.2).}
Averaging leads only at the lenient Acc@0.5 \rev{(\Cref{sec:eval-protocol})}.
\rev{The two synthetic-box rules that assume equal or estimated precision degrade under heterogeneous member precision, while the coordinate-wise median and the medoid do not.}
\rev{The medoid is the only rule among them that returns a box an actual member drew, which keeps each prediction attributable to a single model.}
A Bayesian likelihood-ratio \emph{selection} variant did not help\rev{, over-trusting} agreeing low-error models on shared blind spots.

\paragraph{Member subsets.}
Sweeping every lineup drawn from the balanced members (Appendix D) shows that accuracy is monotone in lineup size and that no member is load-bearing: removals cost between $-1.2$ and $-4.1$ mean, and the Qwen-free pair (Gemma-4 $+$ GLM-4.6V) alone attains 53.2, above the 397B single-model reference.
The benefit therefore stems from \rev{cross-family} diversity and the two fixed rules rather than from access to any one model, in line with the order-independent scaling over an eight-model pool in \Cref{fig:kscaling}.

\section{Conclusion}
\label{sec:conclusion}

We studied adversarial referring expression comprehension, where same-category distractors and negation make one-pass grounding brittle, and showed that training-free test-time fusion of a few frozen MLLMs recovers reliable grounding when their errors are \rev{weakly correlated across families}.
Cross-model agreement supplies a \rev{label-free, agreement-based confidence} signal that single-model confidence \rev{cannot replace}: confidence misses the distractor trap, while different models rarely fall for the same confounder.
Returning the medoid, a real member box, preserves strict-IoU accuracy where coordinate averaging collapses (16.7 to 37.2 Acc@0.9 with a coarse-box member).
With these two fixed rules, \TableOurs{} reaches 55.4 mean on Ref-Adv-s, above every published single model up to the 397B reference, transfers to RefCOCO+, and holds for a balanced lineup with no dominant member (55.3).

\paragraph{Limitations.}
Fusion is regime-dependent and multiplies inference cost.
When one model already subsumes the others, fusion matches but need not exceed it, so the strongest member should be reported alongside; the balanced lineup shows the favorable regime, where the margin over the best member grows from $+3.2$ to $+5.4$ mean (Appendix D).
The localization step is related to precision-weighted box fusion; our contribution is the confidence-free selection rule and the regime analysis rather than a new estimator, and we test REC on only two benchmarks with fixed lineups.

\paragraph{Future work.}
The regime analysis suggests adaptive fusion that consults the weaker members only when the strongest looks suspect, retaining full-lineup accuracy at lower cost.
Being label-free and geometric, the rules could also distil the lineup into one compact model or extend to segmentation and video.


%
%
\bibliographystyle{splncs04}
\bibliography{main}

\begin{thebibliography}{10}
\providecommand{\url}[1]{\texttt{#1}}
\providecommand{\urlprefix}{URL }
\providecommand{\doi}[1]{https://doi.org/#1}

\bibitem{Bai2023QwenVL}
Bai, J., Bai, S., Yang, S., Wang, S., Tan, S., Wang, P., Lin, J., Zhou, C.,
  Zhou, J.: {Qwen-VL}: A versatile vision-language model for understanding,
  localization, text reading, and beyond. arXiv preprint arXiv:2308.12966
  (2023)

\bibitem{Bai2025Qwen3VL}
Bai, S., Cai, Y., Chen, R., et~al.: {Qwen3-VL} technical report. arXiv preprint
  arXiv:2511.21631 (2025)

\bibitem{Bai2025Qwen25VL}
Bai, S., Chen, K., Liu, X., et~al.: {Qwen2.5-VL} technical report. arXiv
  preprint arXiv:2502.13923 (2025)

\bibitem{Bodla2017SoftNMS}
Bodla, N., Singh, B., Chellappa, R., Davis, L.S.: {Soft-NMS}: Improving object
  detection with one line of code. In: ICCV. pp. 5561--5569 (2017)

\bibitem{Chen2023Shikra}
Chen, K., Zhang, Z., Zeng, W., Zhang, R., Zhu, F., Zhao, R.: Shikra: Unleashing
  multimodal {LLM}'s referential dialogue magic. arXiv preprint
  arXiv:2306.15195  (2023)

\bibitem{Chen2020UNITER}
Chen, Y.C., Li, L., Yu, L., El~Kholy, A., Ahmed, F., Gan, Z., Cheng, Y., Liu,
  J.: {UNITER}: Universal image-text representation learning. In: ECCV. pp.
  104--120 (2020)

\bibitem{Dai2025DeRIS}
Dai, M., Cheng, W., Liu, J.j., Yang, S., Cai, W., Sun, Y., Yang, W.: {DeRIS}:
  Decoupling perception and cognition for enhanced referring image segmentation
  through loopback synergy. In: ICCV. pp. 19936--19946 (2025)

\bibitem{dong2026refadv}
Dong, Q., Yang, K., Ju, L., Zhao, H., Zhang, Y., Wang, Y., Zeng, H., Lu, J.,
  Fu, Y.: {Ref-Adv}: Exploring {MLLM} visual reasoning in referring expression
  tasks. In: ICLR (2026)

\bibitem{Du2024Debate}
Du, Y., Li, S., Torralba, A., Tenenbaum, J.B., Mordatch, I.: Improving
  factuality and reasoning in language models through multiagent debate. In:
  ICML (2024)

\bibitem{GLMVTeam2025GLM45V}
{GLM-V Team}: {GLM-4.5V} and {GLM-4.1V-Thinking}: Towards versatile multimodal
  reasoning with scalable reinforcement learning. arXiv preprint
  arXiv:2507.01006 (2025)

\bibitem{Google2026Gemma4}
{Google}: Gemma 4 model overview. \url{https://ai.google.dev/gemma/docs/core}
  (2026), accessed: 2026-06-30

\bibitem{Jiang2023LLMBlender}
Jiang, D., Ren, X., Lin, B.Y.: {LLM-Blender}: Ensembling large language models
  with pairwise ranking and generative fusion. In: ACL. pp. 14165--14178 (2023)

\bibitem{Jiang2026RexThinker}
Jiang, Q., Chen, X., Zeng, Z., Yu, J., Zhang, L.: Rex-thinker: Grounded object
  referring via chain-of-thought reasoning. In: ICLR (2026)

\bibitem{Kamath2021MDETR}
Kamath, A., Singh, M., LeCun, Y., Synnaeve, G., Misra, I., Carion, N.: {MDETR}:
  Modulated detection for end-to-end multi-modal understanding. In: ICCV. pp.
  1780--1790 (2021)

\bibitem{kazemzadeh2014refcoco}
Kazemzadeh, S., Ordonez, V., Matten, M., Berg, T.: Referitgame: Referring to
  objects in photographs of natural scenes. In: EMNLP. pp. 787--798 (2014)

\bibitem{Li2024MoreAgents}
Li, J., Zhang, Q., Yu, Y., Fu, Q., Ye, D.: More agents is all you need. TMLR
  (2024)

\bibitem{Li2022GLIP}
Li, L.H., Zhang, P., Zhang, H., Yang, J., Li, C., Zhong, Y., Wang, L., Yuan,
  L., Zhang, L., Hwang, J.N., Chang, K.W., Gao, J.: Grounded language-image
  pre-training. In: CVPR. pp. 10965--10975 (2022)

\bibitem{Liu2026AIF}
Liu, C., Choi, W., Zhang, C., Oh, T.H.: Aligning what vision-language models
  see and perceive with adaptive information flow. In: CVPR. pp. 24706--24715
  (2026)

\bibitem{Liu2024GroundingDINO}
Liu, S., Zeng, Z., Ren, T., Li, F., Zhang, H., Yang, J., Jiang, Q., Li, C.,
  Yang, J., Su, H., Zhu, J., Zhang, L.: Grounding {DINO}: Marrying {DINO} with
  grounded pre-training for open-set object detection. In: ECCV. pp. 38--55
  (2024)

\bibitem{Mao2016RefExp}
Mao, J., Huang, J., Toshev, A., Camburu, O., Yuille, A.L., Murphy, K.:
  Generation and comprehension of unambiguous object descriptions. In: CVPR.
  pp. 11--20 (2016)

\bibitem{NVIDIA2025Nemotron}
{NVIDIA}: Nemotron nano omni model card. \url{https://huggingface.co/nvidia}
  (2025)

\bibitem{Peng2024Kosmos2}
Peng, Z., Wang, W., Dong, L., Hao, Y., Huang, S., Ma, S., Wei, F.: Grounding
  multimodal large language models to the world. In: ICLR (2024)

\bibitem{Qiao2021Survey}
Qiao, Y., Deng, C., Wu, Q.: Referring expression comprehension: A survey of
  methods and datasets. IEEE TMM  \textbf{23},  4426--4440 (2021)

\bibitem{QwenTeam2026Qwen35}
{Qwen Team}: {Qwen3.5}: Towards native multimodal agents.
  \url{https://qwen.ai/blog?id=qwen3.5} (February 2026)

\bibitem{QwenTeam2026Qwen36A3B}
{Qwen Team}: {Qwen3.6-35B-A3B}: Agentic coding power, now open to all.
  \url{https://qwen.ai/blog?id=qwen3.6-35b-a3b} (April 2026)

\bibitem{Rahmanzadehgervi2024Blind}
Rahmanzadehgervi, P., Bolton, L., Taesiri, M.R., Nguyen, A.T.: Vision language
  models are blind. In: ACCV. pp. 18--34 (2024)

\bibitem{Shanmugam2021TTA}
Shanmugam, D., Blalock, D., Balakrishnan, G., Guttag, J.: Better aggregation in
  test-time augmentation. In: ICCV. pp. 1194--1203 (2021)

\bibitem{Solovyev2021WBF}
Solovyev, R., Wang, W., Gabruseva, T.: Weighted boxes fusion: Ensembling boxes
  from different object detection models. Image Vis. Comput.  \textbf{107},
  104117 (2021)

\bibitem{Tao2026DiG}
Tao, Z., Wang, S., Hua, Y., Cao, H., Xu, L.: {DiG}: Differential grounding for
  enhancing fine-grained perception in multimodal large language model. In:
  CVPR. pp. 1695--1705 (2026)

\bibitem{Wang2025InternVL35}
Wang, W., Gao, Z., Gu, L., et~al.: {InternVL3.5}: Advancing open-source
  multimodal models in versatility, reasoning, and efficiency. arXiv preprint
  arXiv:2508.18265 (2025)

\bibitem{Wang2023SelfConsistency}
Wang, X., Wei, J., Schuurmans, D., Le, Q., Chi, E., Narang, S., Chowdhery, A.,
  Zhou, D.: Self-consistency improves chain of thought reasoning in language
  models. In: ICLR (2023)

\bibitem{Wei2026YoutuVL}
Wei, Z., Li, Y., Kan, Z., et~al.: {Youtu-VL}: Unleashing visual potential via
  unified vision-language supervision. arXiv preprint arXiv:2601.19798 (2026)

\bibitem{Wortsman2022ModelSoups}
Wortsman, M., Ilharco, G., Gadre, S.Y., Roelofs, R., Gontijo-Lopes, R., Morcos,
  A.S., Namkoong, H., Farhadi, A., Carmon, Y., Kornblith, S., Schmidt, L.:
  Model soups: Averaging weights of multiple fine-tuned models improves
  accuracy without increasing inference time. In: ICML (2022)

\bibitem{Wu2026ForeSight}
Wu, Z., Wang, T., Wang, S., Liu, N., Zhang, Y.: See further, think deeper:
  Advancing {VLM}'s reasoning ability with low-level visual cues and
  reflection. In: CVPR. pp. 18870--18880 (2026)

\bibitem{Xiao2025Survey}
Xiao, L., Yang, X., Lan, X., Wang, Y., Xu, C.: Towards visual grounding: A
  survey. IEEE TPAMI  \textbf{48}(3),  2749--2771 (2026)

\bibitem{Xu2024LLaVAUHD}
Xu, R., Yao, Y., Guo, Z., Cui, J., Ni, Z., Ge, C., Chua, T.S., Liu, Z., Sun,
  M.: {LLaVA-UHD}: An {LMM} perceiving any aspect ratio and high-resolution
  images. In: ECCV. pp. 390--406 (2024)

\bibitem{You2024Ferret}
You, H., Zhang, H., Gan, Z., Du, X., Zhang, B., Wang, Z., Cao, L., Chang, S.F.,
  Yang, Y.: Ferret: Refer and ground anything anywhere at any granularity. In:
  ICLR (2024)

\bibitem{Yu2016Context}
Yu, L., Poirson, P., Yang, S., Berg, A.C., Berg, T.L.: Modeling context in
  referring expressions. In: ECCV. pp. 69--85 (2016)

\bibitem{Zheng2026DeepEyes}
Zheng, Z., Yang, M., Hong, J., Zhao, C., Xu, G., Yang, L., Shen, C., Yu, X.:
  {DeepEyes}: Incentivizing ``thinking with images'' via reinforcement
  learning. In: ICLR (2026)

\end{thebibliography}

%
\clearpage
\section*{Supplementary Material}
\appendix

\section{Sensitivity to the Clustering Threshold}
\label{sec:appendix-tau}

\Cref{tab:tau} reports the mean accuracy of \TableOurs{} as the IoU clustering threshold~$\tau$ is swept.
The variation stays within $0.4$ (dominant lineup) and $0.1$ (balanced lineup) on Ref-Adv-s, and within $1.0$ on RefCOCO+ testB, across $\tau\in[0.3,0.7]$, confirming that the single fusion hyperparameter is not tuned to $\tau{=}0.5$.

\begin{table}[t]
  \centering
  \caption{Sensitivity to the IoU clustering threshold~$\tau$ (\TableOurs{} mean accuracy).}
  \label{tab:tau}
  \setlength{\tabcolsep}{6pt}
  \begin{tabular}{lccccc}
    \toprule
    $\tau$ & 0.3 & 0.4 & 0.5 & 0.6 & 0.7 \\
    \midrule
    Ref-Adv-s (dominant)\,$\uparrow$ & 55.3 & 55.1 & 55.4 & 55.4 & 55.0 \\
    Ref-Adv-s (balanced)\,$\uparrow$ & 55.3 & 55.4 & 55.3 & 55.4 & 55.4 \\
    RefCOCO+ testB (dominant)\,$\uparrow$ & 68.6 & 68.7 & 69.2 & 69.1 & 68.2 \\
    \bottomrule
  \end{tabular}
\end{table}

\section{Agreement versus Confidence}
\label{sec:appendix-conf}

Could selection rely on each model's own confidence (self-consistency under resampling) instead of cross-model agreement?
We measure both signals on the balanced lineup, whose Qwen member provides $k{=}8$ stochastic samples ($1{,}070$ instances with valid samples).
Both signals are informative about correctness ($I{=}0.10$ bits for confidence, $0.12$ for agreement), but they fail on different instances.
\Cref{tab:conf-agree} shows the 2$\times$2 breakdown for the Qwen member: when the model is confident yet the other members disagree, the distractor trap, accuracy is only 33.3\%, versus 85.0\% when both \rev{signals} hold.

\begin{table}[t]
  \centering
  \caption{Ref-Adv-s accuracy (\%) by self-consistency and cross-model agreement (balanced lineup, Qwen3.5-27B primary box; $n$ in parentheses).}
  \label{tab:conf-agree}
  \begin{tabular}{lcc}
    \toprule
    & Agreement $\ge 2$ models & Agreement $< 2$ \\
    \midrule
    Confident ($s \ge 7$ samples) & 85.0 (579) & 33.3 (39) \\
    Not confident & 55.8 (360) & 29.3 (92) \\
    \bottomrule
  \end{tabular}
\end{table}

\section{Statistical Testing}
\label{sec:appendix-stats}

We assess significance with a paired bootstrap over the full $1{,}142$-instance split ($10{,}000$ resamples) on the primary mean metric.
\TableOurs{} with the dominant lineup exceeds its best member (Qwen3.6) by $+3.2$ (95\% CI $[1.8,4.5]$, $p<0.001$) and the plain ensemble by $+7.2$ (95\% CI $[6.2,8.2]$, $p<0.001$).
With the balanced lineup the margins are $+5.4$ over the best member (95\% CI $[3.7,7.3]$, $p<0.001$) and $+0.8$ over the plain ensemble (95\% CI $[0.1,1.6]$, $p{=}0.016$), with $+2.7$ at strict Acc@0.9 (95\% CI $[1.1,4.5]$, $p<0.01$).
All confidence intervals exclude zero, so the gains on Ref-Adv-s are unlikely to be sampling noise.

\section{Member Subsets of the Balanced Lineup}
\label{sec:appendix-subsets}

\Cref{tab:balanced-lineup} enumerates every lineup drawn from the three balanced members, fused with the same fixed rules.
Accuracy is monotone in lineup size: every pair beats every single model, and the full trio is best on every metric.
No member is load-bearing: removals cost between $-1.2$ (GLM-4.6V) and $-4.1$ (Gemma-4) mean, and the Qwen-free pair (Gemma-4 $+$ GLM-4.6V) alone attains 53.2, above the 397B-parameter single-model reference (52.6, Table 1 in the main paper).

\begin{table}[tb]
  \centering
  \caption{Member subsets of the balanced lineup on Ref-Adv-s; \checkmark{} marks the members in each lineup. Best per column in bold.}
  \label{tab:balanced-lineup}
  \small
  \setlength{\aboverulesep}{0pt}\setlength{\belowrulesep}{0pt}
  \renewcommand{\arraystretch}{1.15}
  \setlength{\tabcolsep}{5pt}
  \begin{tabular}{ccccccc}
    \toprule
    \multicolumn{3}{c}{Members} & \multicolumn{4}{c}{Accuracy} \\
    \cmidrule(lr){1-3}\cmidrule(lr){4-7}
    Qwen3.5-27B & Gemma-4 & GLM-4.6V & Mean\,$\uparrow$ & @0.5\,$\uparrow$ & @0.75\,$\uparrow$ & @0.9\,$\uparrow$ \\
    \midrule
    \checkmark & & & 49.9 & 64.3 & 52.4 & 32.9 \\
    & \checkmark & & 49.4 & 64.7 & 53.5 & 30.0 \\
    & & \checkmark & 48.3 & 60.0 & 51.4 & 33.5 \\
    \midrule
    & \checkmark & \checkmark & 53.2 & 66.3 & 57.7 & 35.5 \\
    \checkmark & & \checkmark & 51.2 & 63.7 & 54.4 & 35.4 \\
    \checkmark & \checkmark & & 54.1 & 68.4 & 58.6 & 35.4 \\
    \midrule
    \rowcolor{oursblue}
    \checkmark & \checkmark & \checkmark & \textbf{55.3} & \textbf{69.1} & \textbf{59.5} & \textbf{37.2} \\
    \bottomrule
  \end{tabular}
\end{table}

\section{Full Ref-Adv-s Comparison}
\label{sec:appendix-full}

\Cref{tab:leaderboard-full} extends Table 1 in the main paper with all remaining single-MLLM baselines.
Rows for our fusion members (GLM-4.6V, Gemma-4-31B, Qwen3.5-27B) are our greedy runs under the official harness; the published leaderboard entry for Qwen3.5-27B is 51.6.

\begin{table}[t]
  \centering
  \caption{Full Ref-Adv-s comparison (mean of Acc@\{0.5, 0.75, 0.9\}), sorted by mean within each group; best per column in bold.}
  \label{tab:leaderboard-full}
  \small
  \setlength{\aboverulesep}{0pt}\setlength{\belowrulesep}{0pt}
  \renewcommand{\arraystretch}{1.15}
  \setlength{\tabcolsep}{3.5pt}
  \begin{tabular}{lcccc}
    \toprule
    System & Mean\,$\uparrow$ & @0.5\,$\uparrow$ & @0.75\,$\uparrow$ & @0.9\,$\uparrow$ \\
    \midrule
    \groupband{5}{Specialist}
    Grounding DINO & 15.9 & 19.7 & 17.2 & 10.7 \\
    DeRIS-B & 21.1 & 28.3 & 22.2 & 12.8 \\
    \midrule
    \groupband{5}{MLLMs}
    Nemotron-Omni-30B & 21.5 & 51.5 & 9.5 & 3.6 \\
    Youtu-VL & 35.0 & 48.5 & 38.1 & 18.5 \\
    Qwen2.5-VL-72B & 37.4 & 54.0 & 40.1 & 18.0 \\
    InternVL3.5-8B & 38.6 & 52.1 & 41.2 & 22.4 \\
    Qwen3-VL-4B-Thinking & 43.6 & 57.6 & 45.5 & 27.8 \\
    Qwen3-VL-8B-Thinking & 45.0 & 59.5 & 48.2 & 27.3 \\
    GLM-4.6V & 48.3 & 60.0 & 51.4 & 33.5 \\
    Gemma-4-31B & 49.4 & 64.7 & 53.5 & 30.0 \\
    Qwen3.5-27B & 49.9 & 64.3 & 52.4 & 32.9 \\
    Qwen3-VL-32B-Thinking & 50.0 & 65.6 & 52.8 & 31.6 \\
    Qwen3-VL-235B-Thinking & 50.8 & 67.1 & 53.6 & 31.8 \\
    Qwen3.6-35B-A3B & 52.2 & 65.1 & 55.3 & 36.3 \\
    Qwen3.5-122B-A10B & 52.4 & 67.2 & 55.0 & 35.1 \\
    Qwen3.5-397B-A17B-FP8 & 52.6 & 68.0 & 55.6 & 34.2 \\
    \midrule
    \groupband{5}{Test-time fusion}
    Plain ensemble (dominant) & 48.2 & \textbf{70.2} & 57.7 & 16.7 \\
    Plain ensemble (balanced) & 54.5 & 69.4 & \textbf{59.5} & 34.5 \\
    \rowcolor{oursblue}
    \TableOurs{} (balanced) & 55.3 & 69.1 & \textbf{59.5} & \textbf{37.2} \\
    \rowcolor{oursblue}
    \TableOurs{} (dominant) & \textbf{55.4} & 69.8 & 59.2 & \textbf{37.2} \\
    \bottomrule
  \end{tabular}
\end{table}

\section{Additional Qualitative Results}
\label{sec:appendix-qualitative}

\begin{figure}[t]
  \centering
  \includegraphics[width=\linewidth]{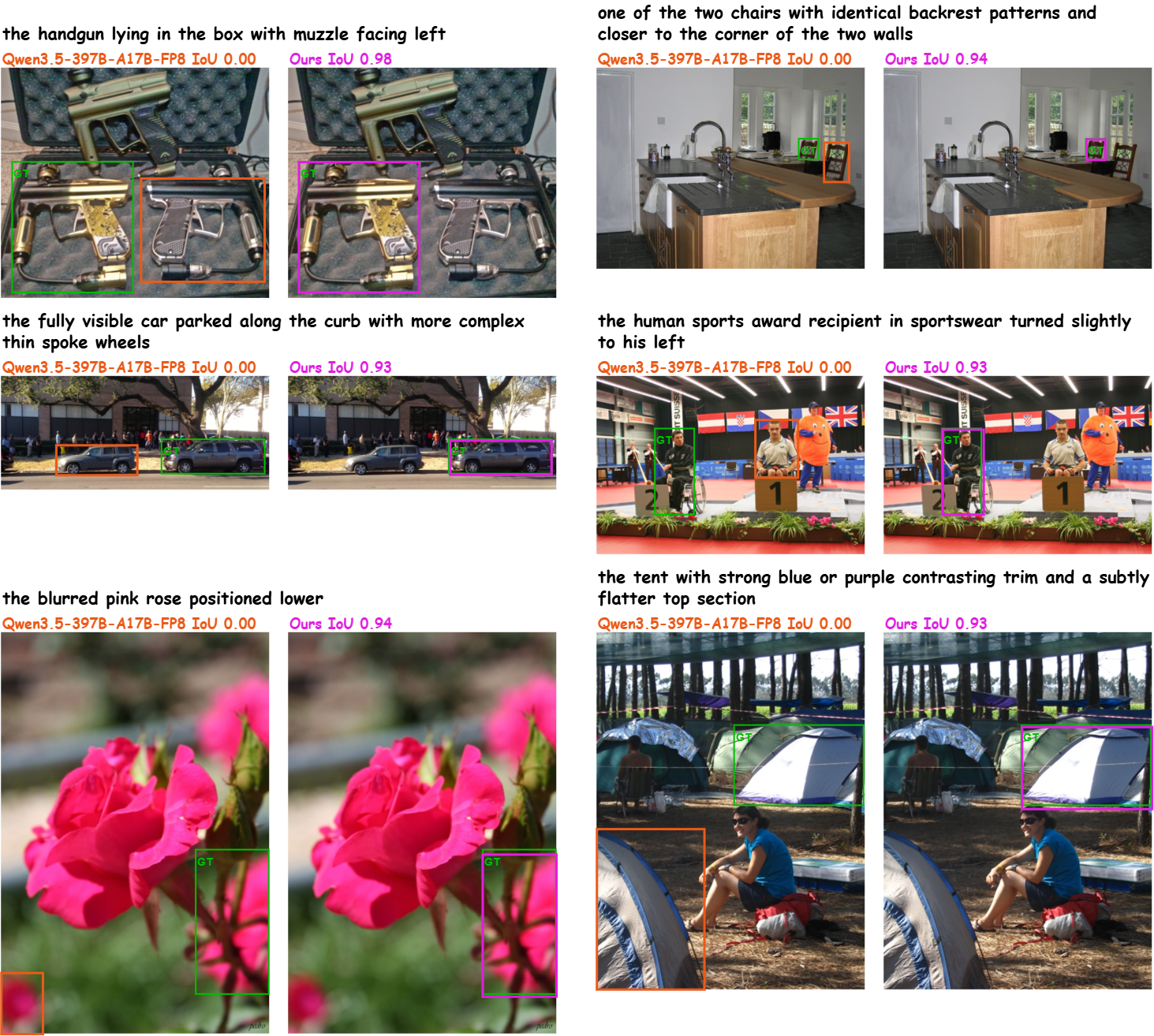}
  \caption{Additional successful Ref-Adv-s examples (companion to Fig. 3 in the main paper).}
  \label{fig:appendix-qual-success}
\end{figure}

\begin{figure}[t]
  \centering
  \includegraphics[width=\linewidth]{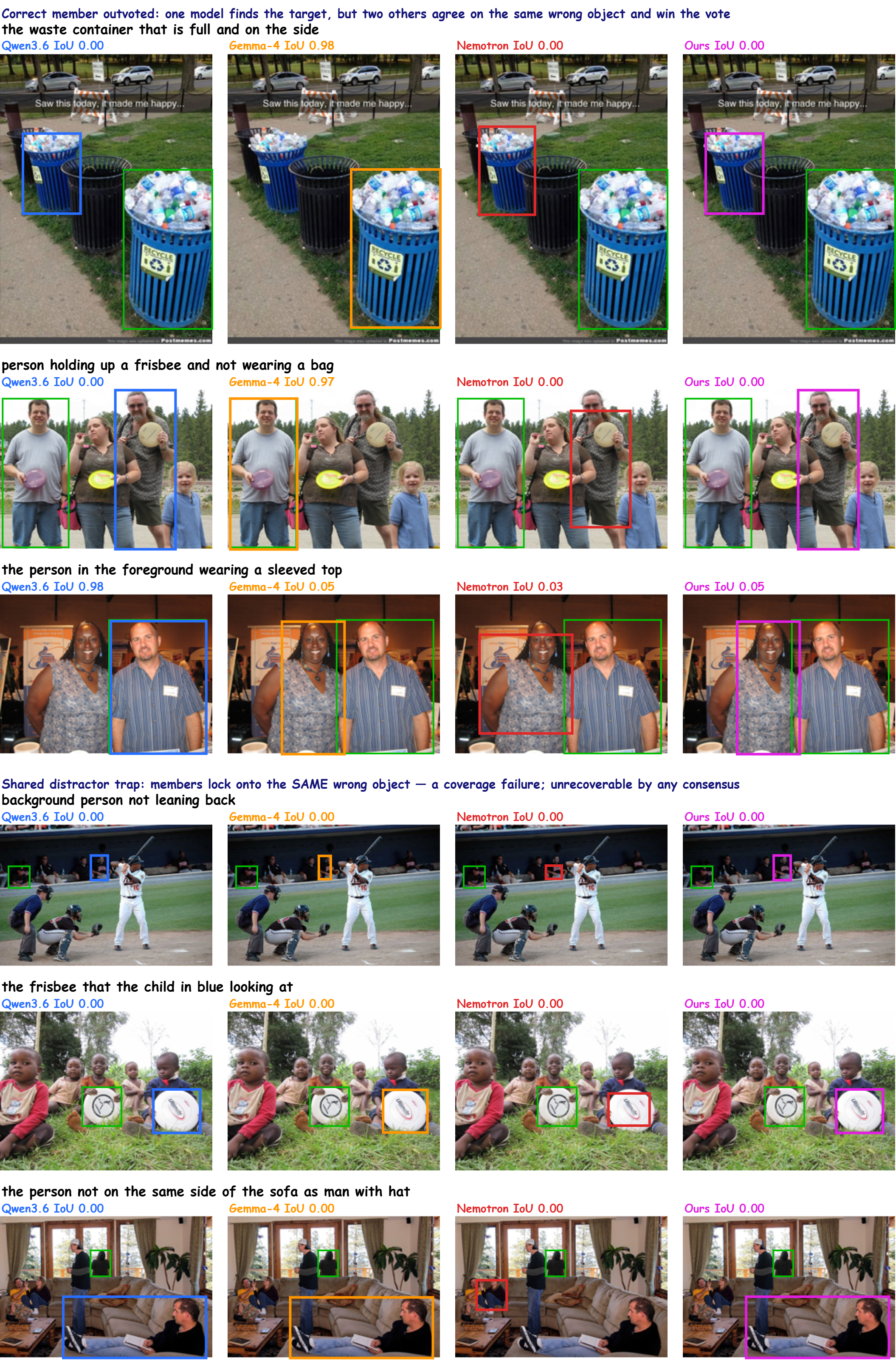}
  \caption{Additional failure modes on Ref-Adv-s (companion to Fig. 4 in the main paper).}
  \label{fig:appendix-qual-failure}
\end{figure}

\Cref{fig:appendix-qual-success} extends Fig. 3 in the main paper with further successful Ref-Adv-s cases in which the largest single-model reference (Qwen3.5-397B-A17B-FP8) misses the target while \TableOurs{} (dominant lineup) grounds it correctly.
\Cref{fig:appendix-qual-failure} extends Fig. 4 in the main paper with further examples of the two failure modes analyzed in Sec. 5.5 of the main paper: a correct member outvoted by two members agreeing on the same wrong object (selection failure), and all members trapped by a shared distractor, which no consensus rule can recover (coverage failure).

\clearpage  
\begin{revblock}
\section{Cross-Family Controls}
\label{sec:appendix-crossfamily}

\Cref{tab:crossfamily} separates family diversity from single-model strength and from the number of predictions.
All rows use the same fixed rules (distinct-model count and medoid, $\tau{=}0.5$) with one primary box per member.
Cross-family fusion (the balanced lineup) reaches 54.6 from an individual mean of 49.2 ($+5.4$), the same-family Qwen trio 52.5 from a higher 50.0 ($+2.5$), with error correlations 0.47 and 0.62, and three samples of the strongest member reach 50.4.
Neither single-model strength nor budget explains the gain; family diversity does.
Because every member contributes one primary box here, the cross-family row reaches 54.6 rather than the 55.3 reported for the balanced lineup in the main paper, which also pools a second, zoomed-in prediction from Gemma-4 and GLM-4.6V.

\begin{table}[!ht]
  \revcolor
  \centering
  \caption{Cross-family controls on Ref-Adv-s ($n{=}1{,}142$, $\tau{=}0.5$, one primary box per member). ``corr.'' is the mean pairwise error correlation among members. Best per column in bold.}
  \label{tab:crossfamily}
  \small
  \setlength{\aboverulesep}{0pt}\setlength{\belowrulesep}{0pt}
  \renewcommand{\arraystretch}{1.15}
  \setlength{\tabcolsep}{4pt}
  \begin{tabular}{lccccc}
    \toprule
    Lineup & Mean\,$\uparrow$ & @0.5\,$\uparrow$ & @0.75\,$\uparrow$ & @0.9\,$\uparrow$ & corr.\,$\downarrow$ \\
    \midrule
    Budget (3$\times$ Qwen3.5-27B) & 50.4 & 65.6 & 53.2 & 32.2 & --- \\
    Same-family (Qwen 9B/27B/35B) & 52.5 & 67.1 & 55.3 & 35.3 & 0.62 \\
    \rowcolor{oursblue}
    Cross-family (\TableOurs{}) & \textbf{54.6} & \textbf{68.9} & \textbf{58.9} & \textbf{36.0} & \textbf{0.47} \\
    \bottomrule
  \end{tabular}
\end{table}

\clearpage
\section{Inference Cost}
\label{sec:appendix-cost}

\Cref{tab:cost} reports FLOPs, memory, and latency for every system, the 397B reference included, on one 8$\times$H100 80\,GB node with vLLM.
FLOPs favor the 397B reference, 19.6 against 29.7 TFLOPs per instance for the dominant lineup ($1.5\times$).
Memory favors \method{}, whose largest member fits one GPU (67 against 378\,GiB): with one GPU per member, three GPUs give 1.4\,s where the 397B needs eight for 1.3\,s, and a single GPU runs the dominant lineup in 2.9\,s where the 397B cannot run.
Rerun through our pipeline, the 397B reference scores 53.7 mean (published 52.6), 1.7 below \method{}.
We therefore make no efficiency claim; the comparison in the main paper is one of accuracy.

\begin{table}[!ht]
  \revcolor
  \centering
  \caption{Inference cost on one 8$\times$H100 80\,GB node (vLLM). TFLOPs per instance; largest checkpoint (GiB; peaks are vLLM preallocation); latency (s) on one GPU and on the node (397B at TP${=}$8, \method{} one GPU per member). The 397B mean is published / rerun through our pipeline. $^\dagger$378\,GiB exceeds one GPU.}
  \label{tab:cost}
  \small
  \setlength{\tabcolsep}{4pt}
  \begin{tabular}{lccccc}
    \toprule
    System & Mean\,$\uparrow$ & TFLOPs\,$\downarrow$ & Ckpt\,$\downarrow$ & 1 GPU\,$\downarrow$ & Node\,$\downarrow$ \\
    \midrule
    Qwen3.5-397B-A17B-FP8 & 52.6 / 53.7 & \textbf{19.6} & 378.3 & ---$^\dagger$ & \textbf{1.3} \\
    \method{} (dominant) & \textbf{55.4} & 29.7 & 67.0 & \textbf{2.9} & 1.4 \\
    \method{} (balanced) & 55.3 & 62.2 & \textbf{58.3} & 5.9 & 3.5 \\
    \bottomrule
  \end{tabular}
\end{table}
\end{revblock}

\end{document}